\documentclass[letterpaper,10pt,journal]{IEEEtran}

\usepackage{graphics} 
\usepackage{epsfig} 
\usepackage{times} 
\usepackage{amsmath} 
\usepackage{amssymb}  

\usepackage{amsthm}
\usepackage{tikz}
\usepackage{graphicx}
\usepackage{xcolor}
\usepackage[font=footnotesize,labelfont=bf]{caption}
\usepackage{subcaption}
\usepackage{algorithm}
\usepackage{algpseudocode}
\algrenewcommand\algorithmicindent{0.5em}
\usepackage{comment}

\usepackage[dvipsnames]{xcolor}
\colorlet{citeblue}{blue!50!black}
\colorlet{linkred}{red!50!black}
\usepackage[colorlinks=true,
            linkcolor=blue,      
            citecolor=blue,      
            urlcolor=blue,       
            bookmarks=true]{hyperref}
\usepackage[isbn=false,
url=true,
doi=false,
backend=biber,
style=ieee,
bibstyle=ieee, 
citestyle=numeric-comp, 
sortcites=true, 
maxbibnames=10,
giveninits=true, 
backref=false]{biblatex}
\usepackage{booktabs}

\newcommand{\argmax}{\arg\max}

\newcommand{\paramspace}{\mathbb{X}}
\newcommand{\param}{\mathbf{x}}

\newcommand{\dataset}{D}

\newtheorem{lemma}{Lemma}

\newtheorem{definition}[lemma]{Definition}
\newtheorem{assumption}[lemma]{Assumption}

\begin{document}

\markboth{IEEE Robotics and Automation Letters. Preprint Version. Accepted February, 2026}
{Menn \MakeLowercase{\textit{et al.}}: Preferential Bayesian Optimization with Crash Feedback} 

\author{Johanna Menn$^{1}$, David Stenger$^{1,2}$, and Sebastian Trimpe$^{1}$%
\thanks{Manuscript received: August 20, 2025; Revised November 19, 2025; Accepted February 11, 2026.}%
\thanks{© 2026 IEEE. Personal use of this material is permitted. Permission from IEEE must be obtained for all other uses, in any current or future media, including reprinting/republishing this material for advertising or promotional purposes, creating new collective works, for resale or redistribution to servers or lists, or reuse of any copyrighted component of this work in other works. DOI: 10.1109/LRA.2026.3665446}%
\thanks{This paper was recommended for publication by Editor J. Kober upon evaluation of the Associate Editor and Reviewers’ comments.
This work was funded in part by the Deutsche Forschungsgemeinschaft (DFG, German Research Foundation) under Germany's Excellence Strategy -- EXC-2023 Internet of Production -- 390621612 and project TR 1433/4-1.
Computations were performed using resources granted by RWTH Aachen University
under Project rwth1705. This work was conducted in part within the Helmholtz
School for Data Science in Life, Earth and Energy (HDS-LEE).}
\thanks{$^{1}$Johanna Menn, David Stenger, and Sebastian Trimpe are with the Institute for Data Science in Mechanical Engineering (DSME),
RWTH Aachen University, Aachen, Germany
(\{johanna.menn, david.stenger, trimpe\}@dsme.rwth-aachen.de).}%
\thanks{$^{2}$David Stenger is with aiXopt GmbH, Aachen, Germany.}%
}
\title{Preferential Bayesian Optimization with Crash Feedback}


\maketitle
\IEEEpubidadjcol

\begin{abstract}
Bayesian optimization is a popular black-box optimization method for parameter learning in control and robotics. It typically requires an objective function that reflects the user's optimization goal. However, in practical applications, this objective function is often inaccessible due to complex or unmeasurable performance metrics. Preferential Bayesian optimization (PBO) overcomes this limitation by leveraging human feedback through pairwise comparisons, eliminating the need for explicit performance quantification. 
When applying PBO to hardware systems, such as in quadcopter control, crashes can cause time-consuming experimental resets, wear and tear, or otherwise undesired outcomes.
Standard PBO methods cannot incorporate feedback from such crashed experiments, resulting in the exploration of parameters that frequently lead to experimental crashes.  
 We thus introduce CrashPBO, a user-friendly mechanism that enables users to both express preferences and report crashes during the optimization process. Benchmarking on synthetic functions shows that this mechanism reduces crashes by 63~\% and increases data efficiency. Through experiments on three robotics platforms, we demonstrate the wide applicability and transferability of CrashPBO, highlighting that it provides a flexible, user-friendly framework for parameter learning with human feedback on preferences and crashes.
 
 Video: \url{https://www.youtube.com/watch?v=2FTbzqT4wP8}
\end{abstract}


\section{INTRODUCTION}
\IEEEPARstart{A}{rtificial} intelligence methods such as Bayesian optimization (BO) efficiently find optimal parameter settings in robotic applications and have demonstrated success in numerous use cases \cite{lechuz2024bayesian, hose2024fine, marco2016automatic}. However, finding optimal parameters typically requires the specification of a numeric objective, which demands a rigorous understanding of optimal system behavior and is a challenge in its own right. In contrast, when humans observe a robotic system, they often readily assess whether a system behaves as desired or not, while they will struggle to suggest optimal parameters in high-dimensional spaces. In this work, we combine BO’s systematic search with human intuition by extending preferential BO (PBO) for optimizing robotic system parameters. PBO \cite{gonzalez2017preferential} circumvents the need for an explicit numerical objective function by inferring task performance directly from human feedback. Because humans find it difficult to provide objective ratings \cite{sadigh2017active}, PBO iteratively proposes parameter pairs, or ``duels'', for testing. This enables human users, or as they are called in PBO, decision makers (DMs), to indicate their preferred outcomes without requiring in-depth knowledge of parameter influence.
\begin{figure}[t]
    \centering
    \includegraphics[width=0.45\textwidth]{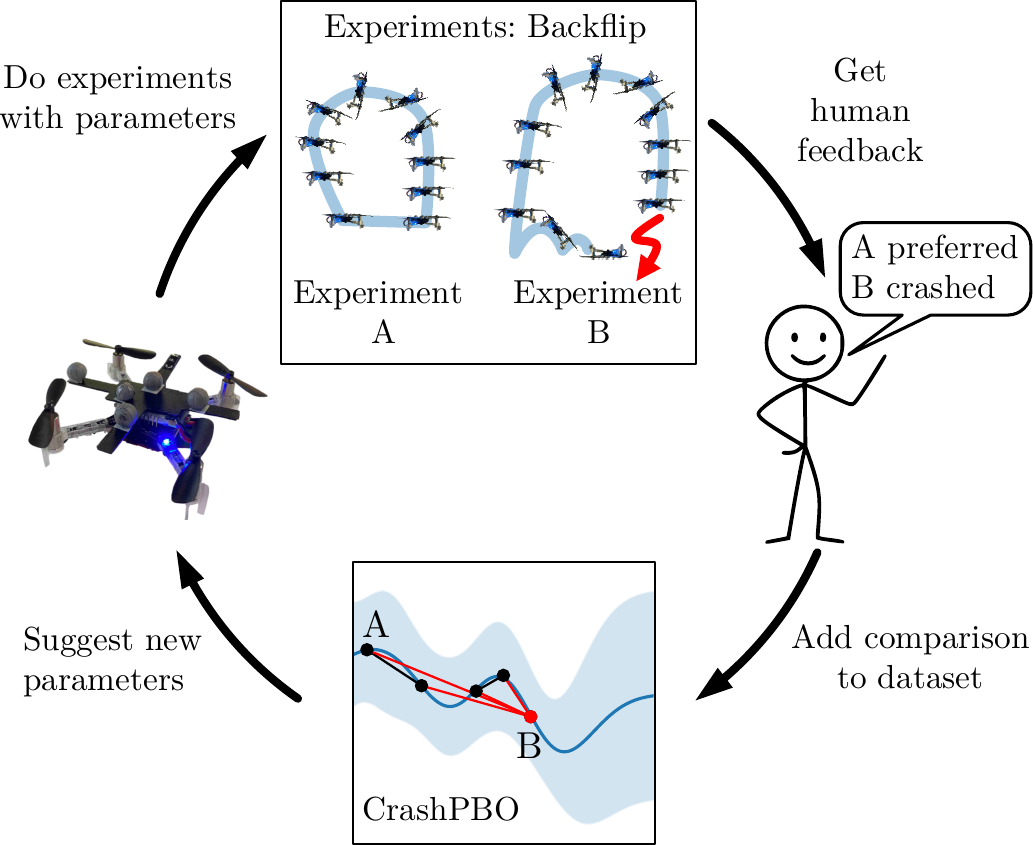}
    \caption{CrashPBO enables optimization of robotic tasks directly from human preferences. The human provides feedback on preferences (``Experiment A was better than B'') and crashes (i.e., totally undesirable experiments). A crashed experiment is ranked worse than all successful ones, preventing the exploration of unsafe or undesirable regions.}
    \label{fig:fig1}
\end{figure}
Despite the advantages of PBO, its application to robotic systems with inherent instability remains limited. Testing specific parameters for these systems can lead to failed experiments (crashes), which slow down optimization, require system resets, and increase mechanical stress. Standard PBO treats such outcomes as \textit{normal} experiments and therefore provides no special signal that helps avoid unsafe or highly undesirable regions. As a result, the algorithm may repeatedly explore risky or undesired settings.

In this work, we introduce CrashPBO (Figure \ref{fig:fig1}), a novel feedback mechanism that allows users to report crashes as outcomes that are strictly worse than all successful experiments. The term ``crash'' follows the concept of crash constraints in BO \cite{marco2021robot, stenger2022benchmark}, an established approach for handling failed evaluations with no measurable objective value. In CrashPBO, a crash does not necessarily correspond to a physical failure, as users may also flag strongly disliked yet technically successful trials as crashes. This preserves the entirely subjective nature of PBO by removing the need to quantify objectives or constraints and by allowing users to reject outcomes without formal definitions.

We validate CrashPBO through empirical studies on synthetic functions, showing that it substantially reduces crashes compared to traditional PBO and does not influence optimization efficiency by incorporating feedback from crashed experiments. Further, we investigate different comparison modes by comparing three strategies for selecting experimental duels. This allows us to identify the most sample-efficient approach for hardware experiments, reducing the number of experiments needed. 

\begin{figure*}[ht]
  \centering
  \begin{subfigure}[t]{0.32\textwidth}
    \centering
    \vspace{0pt}
    \includegraphics[height=3.8cm, trim=7cm 0 7cm 0, clip]{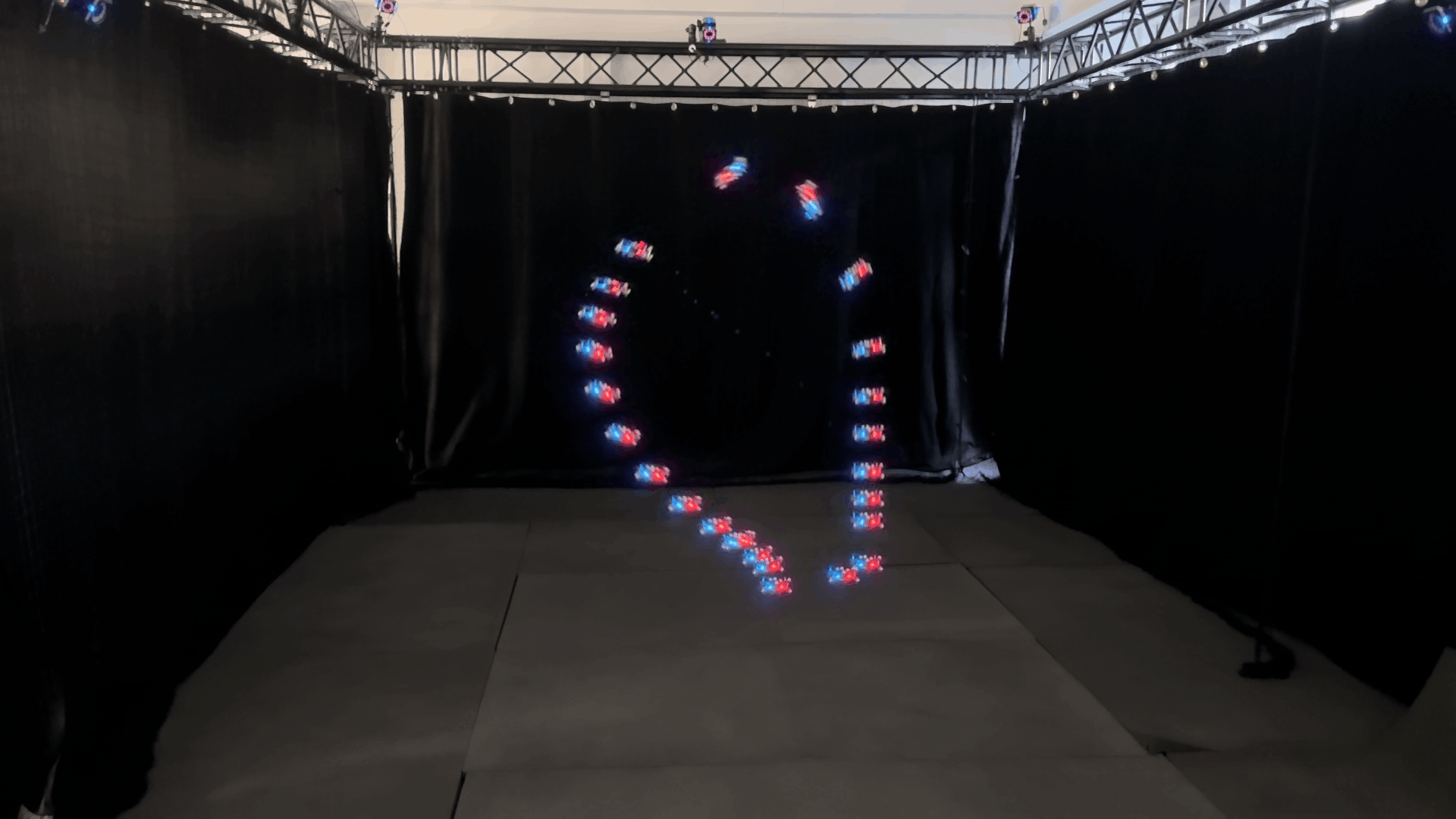}
    \caption{Subjective tuning of quadcopter backflips}
    \label{fig:fig2_drone}
  \end{subfigure}
  \hfill
  \begin{subfigure}[t]{0.32\textwidth}
    \centering
    \vspace{0pt}
    \includegraphics[height=3.8cm, trim=7cm 0 7cm 0, clip]{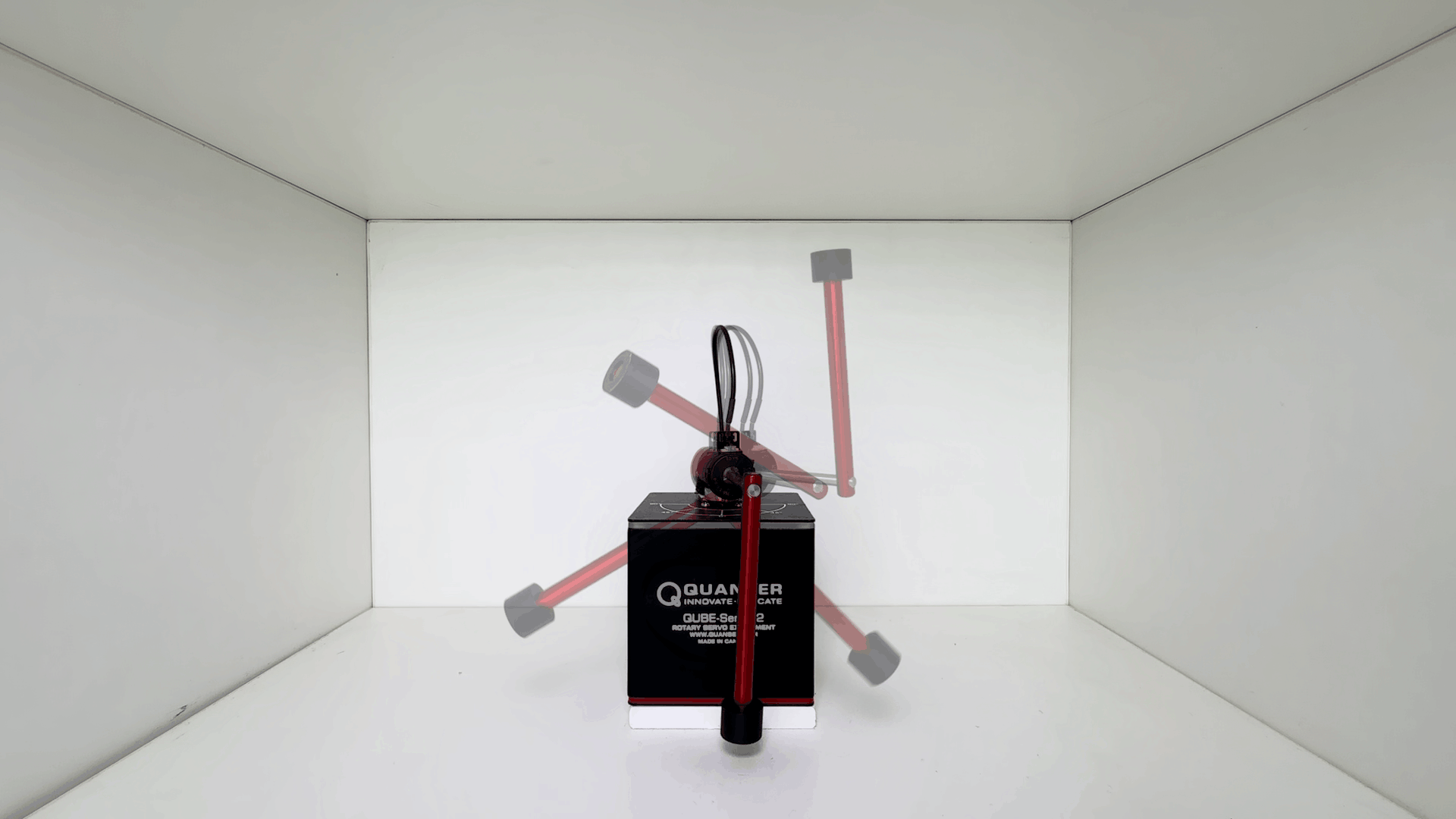}
    \caption{Automatic controller tuning without the need to define a performance function}
    \label{fig:fig2_pendulum}
  \end{subfigure}
  \hfill
  \begin{subfigure}[t]{0.32\textwidth}
    \centering
    \vspace{0pt}
    \includegraphics[height=3.8cm]{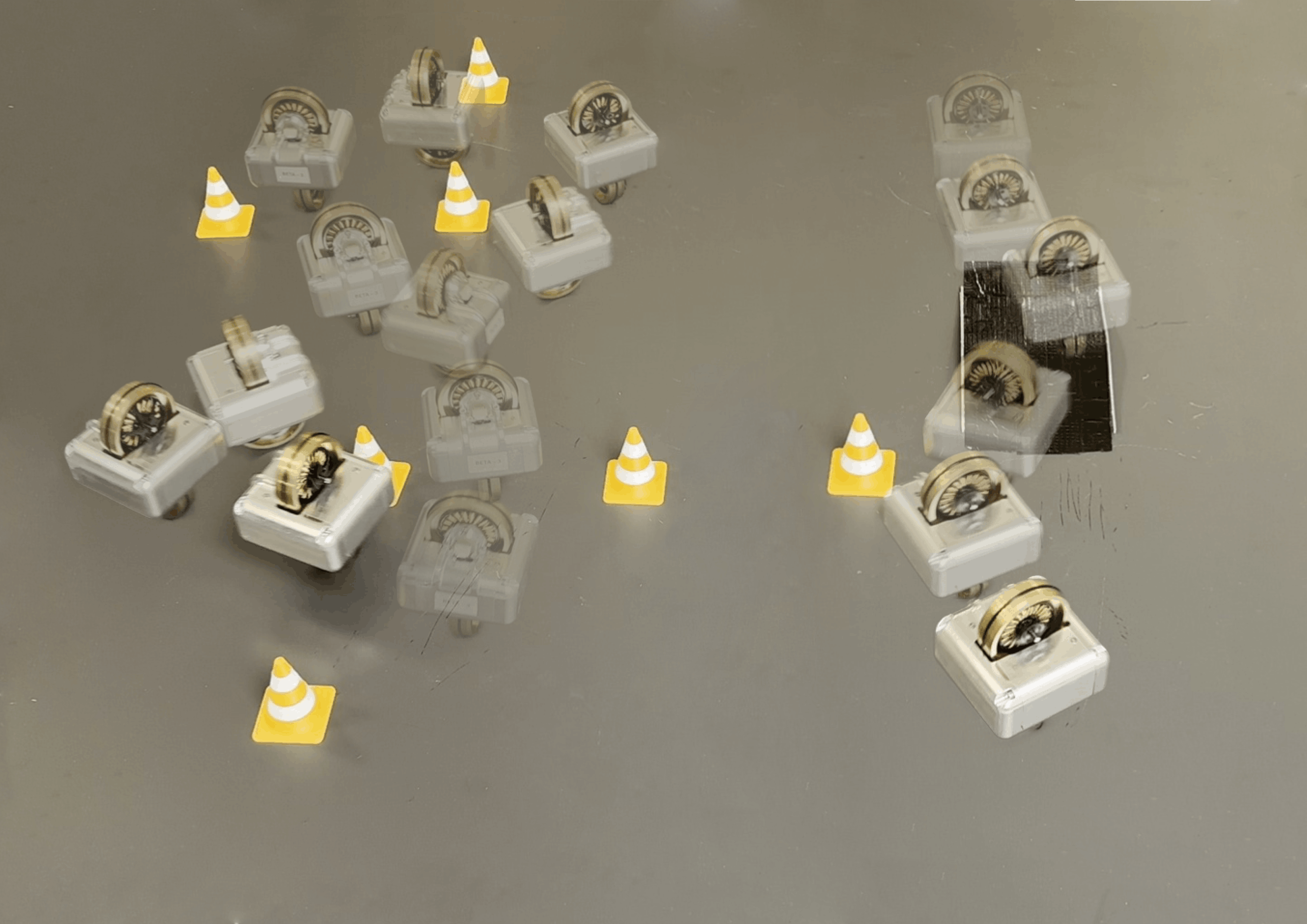}
    \caption{Tuning of control parameters of the Mini Wheelbot for two experiments, figure 8 (left) and ramp (right)}
    \label{fig:fig2_wheelbot}
  \end{subfigure}
  \caption{Applications of CrashPBO}
  \label{fig:three_graphs}
\end{figure*}

To demonstrate the practical applicability of CrashPBO, we perform hardware experiments across three robotic platforms (Figure \ref{fig:three_graphs}): Quadcopter backflips \cite{antal2022nonlinear}, Furuta pendulum swing-up \cite{Astrom2000swinging}, and balancing unicycle robot remote control \cite{hose2025mini}. 
These experiments are new applications for PBO and demonstrate the versatility and fast transferability of CrashPBO. We investigate time savings resulting from fewer experimental resets, reduced wear and tear, and the rejection of subjectively poor parameter sets. As CrashPBO is hyperparameter-free and only influences the data generation, it can be easily applied to new robotic systems and combined with all PBO acquisition functions, foregoing the need to define and shape numeric objective or constraint functions.

In summary, our key contributions are: 
\begin{itemize}
   \item CrashPBO: We propose a hyperparameter-free mechanism to introduce crash feedback into PBO.
    \item Investigation of comparison modalities: We compare the effect of how experimental duels are chosen in PBO on synthetic functions to find the most sample-efficient variant for hardware experiments.
    \item Robot applications: We conduct three hardware experiments that demonstrate the versatility and effectiveness of CrashPBO in real-world robotic systems. 
\end{itemize}
The source code can be accessed at \url{https://github.com/Data-Science-in-Mechanical-Engineering/crashpbo}.

\section{Related Work}
In this work, we propose a mechanism to deal with crashed experiments in PBO, which only requires comparative feedback between two experimental outcomes and binary feedback for crashes.

In standard (non-preferential) BO \cite{garnett2023bayesian}, failed or undesirable experiments are typically addressed through safe BO \cite{sui2015safe}, constrained BO \cite{gardner2014bayesian}, or crash constraints \cite{marco2021robot, stenger2022benchmark}. All these variants have been successfully applied in real-world applications, e.g., safe BO in \cite{berkenkamp2023bayesian, menn2024lipschitz, holzapfel2024event, wang2025safety, wei2024safe}, constrained BO in \cite{khosravi2022safety}, and crash constraints in \cite{marco2021robot, zimmermann2025bayesian}. Safe and constrained BO include undesirable behavior as constraint functions, which are learned from experiments and modeled as Gaussian processes (GP). Safe BO enforces constraint satisfaction in every iteration with high probability and requires regularity assumptions on the constraint functions to achieve safe exploration \cite{sui2015safe, berkenkamp2016safe, bottero2022information, fiedler2024on}. Constrained BO \cite{gardner2014bayesian} only enforces feasibility for the final optimizer, allowing temporary violations and global exploration without safety guarantees. 
Crash constraints \cite{marco2021robot, stenger2022benchmark, von2024local} do not require a measurable constraint function, but only binary feedback, if the experiment failed. This is incorporated in BO by an additional GP to classify crash and safe regions \cite{marco2021robot}, or by virtual data points to steer the optimization away from the crashed region \cite{stenger2022benchmark}. 
 Our approach is similar to the virtual data point approach \cite{stenger2022benchmark}, as we identify crashed points as significantly worse than safe evaluations through additional virtual comparisons. This form of crash feedback is new to \emph{preferential} BO. While it does not guarantee safety, it makes PBO suitable for domains where experiments can fail or yield subjectively undesirable outcomes, without the need to formalize and measure objective and constraint functions. 

 PBO (without crash constraints) has been used in multiple applications to tune the parameters of robotic systems with human interactions. Different PBO algorithms have been proposed for human gait optimization in \cite{tucker2020preference,tucker2020human} and for quadrupedal robots in \cite{cosner2022safety}, which includes safety-aware behavior in PBO. In \cite{kupcsik2015learning}, a learning framework for robot-to-human object handover is proposed. The human can give either preference feedback or an absolute evaluation of the outcome. Moreover, the potential of PBO for controller tuning in simulations was investigated in \cite{coutinho2024human}, demonstrating its ability to speed up the controller tuning process when no quantifiable target function is available. However, all these studies lack a mechanism for rejecting unwanted parameters, which motivates the introduction of a crash mechanism that can be integrated into other PBO frameworks. Additionally, we present three new applications of PBO.  

Other non-BO preference learning frameworks like GLISp \cite{bemporad2021global} and C-GLISp \cite{zhu2021c} have been used for MPC tuning with human preferences \cite{zhu2021preference, zhu2021c} in simulation. GLISp has been applied to different robot-human interaction use cases like \cite{campagna2024promoting} and \cite{roveda2023optimal}.
In settings similar to our work, C-GLISp permits users to state whether the experiment’s performance is satisfactory. To utilize this feedback, C-GLISp introduces a constraint function, which is learned in addition to the preference function. In contrast, we use the feedback directly instead of learning an additional constraint function, and CrashPBO can be combined with any PBO acquisition function, as only data generation is influenced.  

\section{Problem formulation}
We aim to maximize the unknown performance function  $f: \paramspace \rightarrow \mathbb{R}$, where the parameter space is $\paramspace \subseteq \mathbb{R}^d$ and $d \in \mathbb{N}$ is the number of input dimensions. 
However, we cannot observe the function directly. We can only observe $k = 0, \ldots, K$ pairwise comparisons (duels) between the experimental outcomes of the parameters of $\param_{\mathrm{A},k}$ and $\param_{\mathrm{B},k}$.
This can be formalized as the $\{0,1\}$-valued random variable $\pi_k$ such that
\begin{equation}
        \pi_k = \begin{cases}
        0, & \text{if $y_{A,k} > y_{B,k}$},\\
        1, & \text{otherwise},
    \end{cases}
\label{eq:pi}
\end{equation}
where $\param_{i,k} \in \paramspace$ and $y_{i,k}$ is
\begin{equation}
    y_{i,k} = f(\param_{i,k}) + \epsilon_{i,k} , \quad i \in \{0,1\},
    \label{eq:noise}
\end{equation}
with ${\epsilon \sim \mathcal{N}(0, \sigma^2)}$ being independent and identically distributed Gaussian noise.
The data set \mbox{$\dataset = \{(\param_{\mathrm{A},k}, \param_{\mathrm{B},k}, \pi_k)\}_{k = 0, \ldots, K}$} is created by observing $\pi_k$ $(K+1)$ times. Typically, this corresponds to a human providing preferences $\pi_k$ for each $k$.

As we consider unknown constraints, we treat feedback about unwanted behavior, including hardware crashes, as constraint information. 

\begin{definition}
A crash is any user-defined unwanted behavior that may require a manual or automatic reset but does not render the system unrecoverable.
\end{definition}

Therefore, we introduce the unknown \emph{satisfaction set}, where no crashes occur $\Omega_s \subseteq \mathbb{X}$. We assume that the DM can directly evaluate an experimental outcome as ``satisfying'' or ``crashed'' and formalize this as the satisfaction function $S:\paramspace\rightarrow \{ 0, 1 \}$ with 
\begin{equation}
\label{eq:crash}
    S(\param)= \begin{cases}
    0, & \text{if $\param \notin \Omega_s$},\\
    1, & \text{otherwise}.
\end{cases}
\end{equation}

We aim to find the maximizer of the unknown function $f$ that satisfies the satisfaction function $S$. During the optimization, experiments can crash, and we only evaluate comparisons and the satisfaction function. The constrained optimization problem is 
\begin{equation}
\label{eq:problem}
    \begin{aligned}
        & \max_{\param \in \mathbb{X}} \quad f(\param) \\
        & \quad \text{s.t.} \quad S(\param) = 1.
    \end{aligned}
\end{equation}

\begin{assumption}
\label{ass:2}
We assume that the initial comparison contains at least one parameter vector $x$ that fulfills $S(x) = 1$.  
\end{assumption}
This assumption is necessary, as two crashed points in the first comparison would render the comparison impossible, and it corresponds to the assumption of a feasible initial point in standard crash constraints \cite{von2024local}.

BO with preferential feedback \eqref{eq:pi} is standard in PBO. Prior work in PBO has not considered the formulation in \eqref{eq:problem}, leveraging crash information \eqref{eq:crash} in addition. 

\section{Preferential Bayesian Optimization}
BO \cite{garnett2023bayesian} is a sample-efficient black-box optimization algorithm for expensive-to-evaluate and noisy functions. In standard BO \cite{garnett2023bayesian}, the unknown function can be queried directly to get a noisy function value. A GP is typically used as a surrogate model for the function, and an acquisition function suggests the next query. The acquisition function leverages the uncertainty and mean prediction of the GP model to suggest a few queries to reach the function's optimum. 

In this work, we use PBO, a variant of BO specifically designed for settings where only pairwise comparisons can be observed as described in \eqref{eq:pi}.  
Bayesian models best account for this by introducing an appropriate likelihood function connecting $f$ and the comparison measurements. The resulting posterior distribution is known as a pairwise GP \cite{chu2005preference}, and constitutes our model. The acquisition functions in PBO differ from standard BO, as the uncertainty of the function remains high at the queried points without direct measurements.  

\subsection{Pairwise Gaussian Processes}
GPs are a class of non-parametric models and can be used to model functions on arbitrary input domains. 
A GP $f$ is uniquely defined by its mean function $m(\param)$ and the covariance function $k_{xx}(\param, \param')$. Further, we assume, as is common in literature, a zero prior mean function, such that $f \sim \mathcal{GP}(0, k_{xx})$.

The Gaussian prior over $f$ at $K$ training points is
\begin{equation}
    P(\boldsymbol{f}) = \frac{1}{(2\pi)^{K/2} |\Sigma|^{1/2}} \exp{\left( -\frac{1}{2}\mathbf{f}^T \Sigma^{-1} \mathbf{f} \right)}
    \label{eq:prior}
\end{equation}
where $\mathbf{f}=[f(\param_1), f(\param_2), \ldots , f(\param_K)]^\mathrm{T}$ and $\Sigma \in \mathbb{R}^{K \times K}$ is the covariance matrix, whose $ij$-th element is the covariance function $k_{xx}(\param_i,\param_j)$. 

For the likelihood model of the pairwise GP $P(D|\mathbf{f})$, we assume feedback corrupted by i.i.d Gaussian noise \mbox{$y(\param_k)=f(\param_k)+\epsilon_k$}, where $\epsilon \sim \mathcal{N}(0, \sigma^2)$. The probability that the outcome of parameter $\param_{0, k}$ is preferred over $\param_{\mathrm{B},k}$ then is

\begin{equation}
\begin{aligned}
    P(\pi(\param_{\mathrm{A},k}, \param_{\mathrm{B},k}) = 0 | \mathbf{f}) &= P(y_{\mathrm{A},k} > y_{\mathrm{B},k}| f(\param_{\mathrm{A},k}), f(\param_{\mathrm{B},k})) \\
    & = \Phi \left[ \frac{f(\param_{\mathrm{A},k})-f(\param_{\mathrm{B},k})}{\sqrt{2}\sigma} \right],
\end{aligned}
\end{equation}
where $\Phi$ is a standard normal cumulative distribution function. Thus, the formulation for the likelihood is  
\begin{equation}
    P(D|\mathbf{f}) = \prod_{k=1}^N \Phi \left[ \frac{f(\param_{\mathrm{A},k})-f(\param_{\mathrm{B},k})}{\sqrt{2}\sigma} \right].
    \label{eq:likelihood}
\end{equation}
This likelihood model relates binary observations to continuous latent function values and is known as binomial-probit regression in statistics. If, instead of Gaussian noise, Gumbel noise is assumed, logit likelihood models result. For more details, refer to \cite{benavoli2024tutorial}.

The posterior probability is
\begin{equation}
P(\mathbf{f}|D) \propto P(D|\mathbf{f})P(\mathbf{f}).
\label{eq:posterior}
\end{equation}
with the prior defined in \eqref{eq:prior} and the likelihood in \eqref{eq:likelihood}.
The posterior can be estimated via Laplace approximation as a multivariate Gaussian distribution. Details on this can be found in \cite{chu2005preference, brochu2010interactive}.

\subsection{Acquisition Functions and Comparison Modalities}
There are different acquisition functions for PBO. 
In this work, we build upon the commonly used \cite{coutinho2024human} acquisition function EUBO \cite{lin2022preference, astudillo2023qeubo}, which is implemented in the state-of-the-art library BoTorch \cite{balandat2020botorch}. This acquisition function chooses the duel of parameters $(\param_\mathrm{A}, \param_\mathrm{B})$ such that it would maximize the expected utility of the best option, if it were the last to be chosen 
\begin{equation}
    \mathrm{EUBO}(\param_\mathrm{A}, \param_\mathrm{B}) = \mathbb{E}[\max \{f(\param_\mathrm{A}), f(\param_\mathrm{B})\}].
\end{equation}
The acquisition function can be either used to generate both parameters of the duel $(\param_\mathrm{A}, \param_\mathrm{B})$ or only one of the parameters. If only one parameter is newly generated, the other parameter can be either the last observed experimental outcome $\param_\mathrm{last}$, or the best observed experimental outcome so far $\param_\mathrm{best}$. In this work, we compare the effectiveness of these modalities.

\section{Method: Crash Mechanism PBO}

\begin{figure}
    \centering
    \begin{tikzpicture}
        \node[anchor=south west,inner sep=0] (image1) at (0,0) {\includegraphics[width=0.98\linewidth]{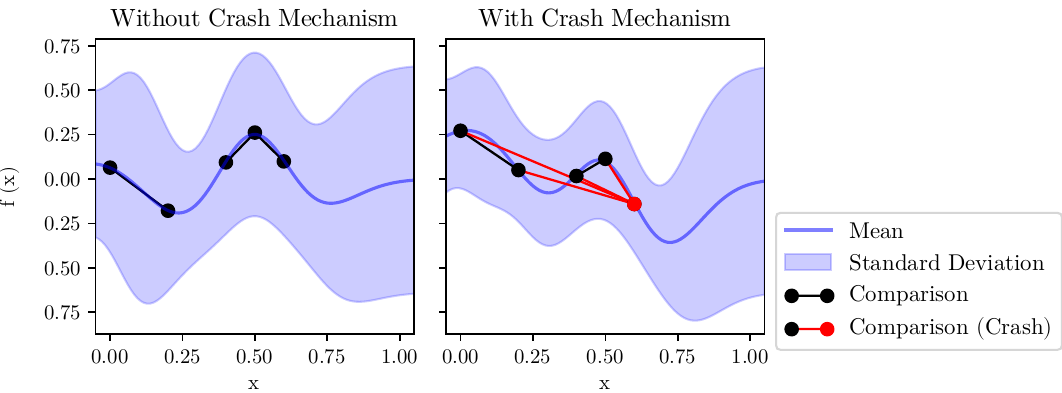}};
        \begin{scope}[x={(image1.south east)},y={(image1.north west)}]
            \node[fill=white,anchor=north east,inner sep=2pt] at (0.97,0.92) {
                \scriptsize
                \begin{tabular}{c}
                    \hline
                    Comparisons\\
                    \hline
                    $0 \succ 0.2$ \\
                    $0.5 \succ 0.4$ \\
                    $0.5 \succ$ \textcolor{red}{$0.6$} \\
                    \hline
                \end{tabular}
            };
        \end{scope} 
    \end{tikzpicture}
    \caption{Crash mechanism in PBO: By adding virtual comparisons of the crashed experiment (red) with all successful experiments (black), we ensure that the posterior is worse in the crashed regions. This reduces the likelihood of exploration in the crashed region. Comparisons between data points are indicated by $\param_\mathrm{A} \succ \param_\mathrm{B}$, meaning that $\param_\mathrm{A}$ is preferred over $\param_\mathrm{B}$, a comparison is indicated with a line in the plot.}
    \label{fig:preferences}
\end{figure}

This section introduces our crash feedback mechanism for PBO, CrashPBO.
The optimization algorithm is described in Algorithm \ref{alg:generalPBO}, and the feedback mechanism for obtaining new data is in Algorithm \ref{alg:feedback}. A comparison between PBO with and without the crash mechanism is illustrated in Figure \ref{fig:preferences}. 

Before starting the optimization loop in Algorithm \ref{alg:generalPBO}, we require an initial data set of comparisons (Require, Algorithm \ref{alg:generalPBO}), just as in standard PBO. Then, the optimization is repeated $T$ times, which is our experiment budget.  In each iteration, we first train a pairwise GP model using the current dataset (Line 2, Algorithm \ref{alg:generalPBO}) as described in \eqref{eq:posterior}. We then use the EUBO acquisition function to suggest a new pair of parameters (Line 3, Algorithm \ref{alg:generalPBO}). Any PBO acquisition function could be used here. 
For the suggested parameters, we obtain the DM's feedback about the outcome of the satisfaction function for each parameter and the result of the preference function (Line 4, Algorithm \ref{alg:generalPBO}). In standard PBO, only the preference feedback is obtained at this point. Then, we augment the dataset (Line 5, Algorithm \ref{alg:generalPBO}). Instead of only adding the outcome of the preference function, we use the feedback mechanism described in Algorithm \ref{alg:feedback} to generate the additional dataset $D_{\mathrm{add}}$. 
Finally, we augment the dataset with the additional dataset (Line 6, Algorithm \ref{alg:generalPBO}) and restart the optimization loop. 

\begin{algorithm}[t]
\caption{PBO with Crash Feedback}
\label{alg:generalPBO}
\begin{algorithmic}[1]
\Require Initial dataset \mbox{$\dataset_0 = \{(\param_{\mathrm{A},n}, \param_{\mathrm{B},n}, \pi_n)\}_{n=1}^N$} 
\For{$t = 1,2,\ldots T$}
    \State Fit a pairwise GP to $D_{t-1}$
    \State Next duel: $\param_{\mathrm{A},t}, \param_{\mathrm{B},t} = \argmax \mathrm{EUBO}(\param_\mathrm{A}, \param_\mathrm{B})$
    \State \mbox{Get DM feedback $S(\param_{\mathrm{A},t}), S(\param_{\mathrm{B},t}), \pi_t$} 
    \State Obtain $\dataset_{\mathrm{add}}$ \Comment{Algorithm 2}
    \State Augment $\dataset_{t} \leftarrow \{ \dataset_{t-1} \cup \dataset_{\mathrm{add}} \}$
\EndFor
\end{algorithmic}
\end{algorithm}

Intuitively, our crash feedback mechanism adds comparisons to the dataset, such that parameters that lead to experimental crashes are always rated worse than those that do not crash. 
To achieve this algorithmically, we track the set of crashed parameters as $X_c \subset \mathbb{X}$ and the set of non-crashed parameters as $X_s \subset \mathbb{X}$. These sets are initialized based on the first comparison, which must contain at least one feasible point, such that $X_s$ is not empty. When a new duel and the corresponding feedback are given, we augment the dataset by using the following rules:

(1) For each crashed experiment in the duel: add virtual preference observations expressing that the crashed point is disadvantaged to each of the non-crashed points in $X_s$ (Lines 3-5, Algorithm 2). Then, add the point to the list of crashed points $X_c$ (Line 5, Algorithm 2).

(2) For each successful experiment in the duel: add virtual preference observations expressing that the successful point is preferred to each of the non-crashed points in $X_c$ (Lines 8-10 in Algorithm 2). Then, add the point to the list of non-crashed points $X_s$ (Line 11, Algorithm 2).

(3) If both parameters are feasible, we also add the comparison of the duel, as we would normally do in PBO (Line 15, Algorithm 2). 

 To reduce crashes, our algorithm adds $\mathcal{O}(|X_s| \cdot |X_c|)$ comparisons. This increases the GP training cost. However, as in BO, we often deal with a low-sample setting and expensive experiments (see Sec. \ref{sec:hexperiment} B), this is often negligible compared to the duration of the experiments.

\begin{algorithm}[h]
\caption{Augment data set, including information on crashes}
\label{alg:feedback}
\begin{algorithmic}[1]
\Require \mbox{current duel $\param_{\mathrm{A}, t}, \param_{\mathrm{B},t}$},  \mbox{set of non-crashed parameters $X_s$} \mbox{set of crashed parameters $X_c$}
\For{$\param_j \in \{\param_{\mathrm{A}, t}, \param_{\mathrm{B}, t}\}$} 
    \If{$S(x_j) = 0$} \Comment{handling of crashed point}
        \ForAll{$\param_s$ in $X_s$} 
            \State $\dataset_{\mathrm{add}} \leftarrow \dataset_{\mathrm{add}} \cup \{ (\param_{j}, \param_{s}, 1) \}$ 
        \EndFor
        \State $X_c \leftarrow X_c \cup \{ \param_j \}$ 
    \ElsIf{$S(\param_j) = 1$} \Comment{handling of non-crashed point}
        \For{$\param_c \in X_c$} 
            \State $\dataset_{\mathrm{add}} \leftarrow \dataset_{\mathrm{add}} \cup \{ (\param_{j}, \param_{c}, 0) \}$  
        \EndFor
        \State $X_s \leftarrow X_s \cup \{ \param_j \}$ 
    \EndIf
\EndFor

\If{$\param_{\mathrm{A}, t} \in X_s$ \textbf{ and } $\param_{\mathrm{B}, t} \in X_s$} \Comment{standard PBO}
    \State $\dataset_{\mathrm{add}} \leftarrow \dataset_{\mathrm{add}} \cup \{ (\param_{\mathrm{A},t}, \param_{\mathrm{B},t}, \pi_t) \}$ 
\EndIf

\end{algorithmic}
\end{algorithm}
\section{Benchmarking on Synthetic Functions}
In this section, we evaluate our method CrashPBO, and different comparison modalities using synthetic test functions and GP sample paths. With this extensive synthetic benchmarking, we select the best setting for CrashPBO to apply to real experiments.

\subsection{Setup}
We evaluate our crash mechanism in combination with EUBO, which we denote as CrashPBO in the experiments and compare to standard EUBO \cite{lin2022preference, astudillo2023qeubo} (without crash constraints). As further baselines, we consider random sampling, max-value entropy search (MES) \cite{wang2017max} as a standard BO algorithm, and SafeOpt \cite{sui2015safe} as a safe BO variant. 
We utilize the implementation from \cite{bottero2022information} and apply the original SafeOpt for low-dimensional problems, while using LineSafeOpt \cite{kirschner2019adaptive} for dimensions higher than three. For all algorithms except safe BO, we use the standard BoTorch \cite{balandat2020botorch} setting for the GP and either fix the hyperparameters for the within-model comparison or use the standard prior. For safe BO, we need to select fixed GP hyperparameters and an upper bound of the Lipschitz constant. We set the parameter that scales the uncertainty bounds to $\beta = 2$, as is common in the literature~\cite{berkenkamp2016safe}.

We use the standard test functions Ackley 2D, Branin 2D, Hartmann 6D function, as well as a Cosine 8D function as implemented in \cite{balandat2020botorch}. Furthermore, we sample paths from a GP with a squared-exponential kernel with lengthscale $l=0.3$ and variance $\sigma_k^2 = 1$ for one to eight dimensions. For each dimension, we sample ten different functions from the GP. For the GP hyperparameters, we use correctly specified GP hyperparameters in the BO algorithms (within-model comparison). We consider the noise setting as described in \eqref{eq:noise} and use Gaussian noise with a variance of $\sigma^2 = 0.01$. We generate comparisons by comparing the noisy function evaluations as described in \eqref{eq:pi} for CrashPBO and EUBO. If the evaluation of the test function is below a certain threshold, CrashPBO receives crash information, while EUBO obtains the comparison for the noisy function values. The non-preferential algorithms, MES, SafeOpt, and random search, obtain the noisy function evaluations directly.

To simulate crashed experiments, we approximate a threshold below which a sample is considered infeasible by gridding the domain and choosing the threshold so that half of the points seen when gridding the functions are below it. This means that approximately half of the domain is infeasible. For each synthetic function and for each sampled GP function, we run the algorithms from 20 different initializations. Given Assumption \ref{ass:2}, this initialization always includes one feasible point.  

When applying PBO on hardware, there are different ways to make comparisons. The standard setting is to compare two new parameters, corresponding to $q=2$ in EUBO, such that for every comparison, two new experiments are required. However, given that there might be only one system on which the experiments can be run, it can make sense to only compare to the best seen parameter (``compare to best''). This is possible if the outcome can be plotted so that the DM can remember the best point, or this point can be queried again. In the ``two new'' case, no explicit notion of a ``best-so-far'' parameter exists, but it is also not needed as only newly proposed pairs are compared. Finally, we can also compare only to the last queried parameters as a direct comparison. The ``compare to best'' and ``compare to last'' settings can be represented in EUBO by setting $q=1$ and using the parameter that won past comparisons or the last used parameter set for the other experiment.

\subsection{Results} 
The results of the synthetic experiments are summarized in Table I. We compare the different settings for EUBO and CrashPBO. In our evaluation, we have two metrics, first, the normalized final performance over all test functions after $10 \times d$ experiments, where $d$ is the dimension of the test function. For this metric, higher performance is better. As a second metric, we use the average number of crashes, so where $S(\param) = 0$, per experiment over all test functions. Here, fewer crashes are better. 
Table \ref{tab:results} shows that the crash feedback reduces the number of crashes and slightly improves the performance independently of the used comparison mode. 
The analysis of the different comparison modes shows that comparing to the best seen parameter leads to the best performance and the lowest number of crashes. 
The comparison to the last experiment increases the number of crashes drastically and reduces performance. The setting with two new experiments is slightly worse than comparing to the best, but leads to fewer crashes in EUBO. 

Based on these results, we choose the setting to compare always to the best observed result so far. This requires some memory of this parametrization. However, this might be available by, e.g., a plot of the experiment.

\begin{figure}
    \centering
    \includegraphics[width=0.49\textwidth]{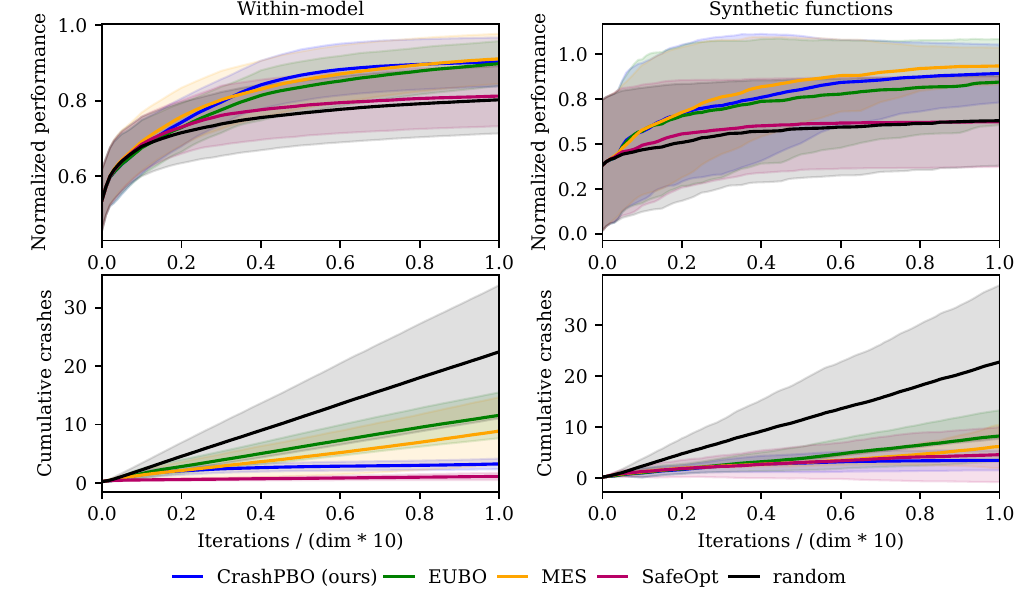}
    
    \caption{Synthetic results showing average normalized performance and average crashes for within-model comparison and synthetic functions. We compare CrashPBO to the PBO method EUBO, the standard BO method MES, the safe BO method SafeOpt, and random search. CrashPBO performs similarly to MES and EUBO and outperforms safe BO and random search while reducing crashes compared to EUBO in both settings.}
    \label{fig:enter-label}
\end{figure}

Figure~\ref{fig:enter-label} separates the within-model comparison from the synthetic-function results.
In the within-model comparison setting, where all GP hyperparameters are correctly specified, CrashPBO performs on par with MES and EUBO while yielding fewer crashed experiments. SafeOpt achieves low crash counts in this idealized setting but performs no better than random search due to its restricted exploration.

On the synthetic test functions, where model mismatch is present, CrashPBO again delivers competitive performance compared to MES and EUBO, and results in slightly fewer crashes than SafeOpt. SafeOpt occasionally stalls in its initial safe region or fails to reach high-performing areas, and its crashes arise from the heuristic GP hyperparameter selection required to run it in practice. A detailed discussion on this can be found in \cite{fiedler2024on}.

CrashPBO and EUBO are competitive with the classical BO methods in terms of performance. Overall, we can conclude that the crash feedback in PBO can reduce the number of crashes by up to 63 \% while achieving, on average, slightly better performance with the same number of experiments. Thus, assuming that a crash experiment takes longer due to the increased effort in the environmental reset, e.g., in \ref{sec:quad}, CrashPBO can reduce overall experimentation time.

\begin{table}[ht]
\centering
\caption{Comparison on average performance (``higher is better'', $\uparrow$) and crashes (``fewer is better'', $\downarrow$) over four test functions and within-model comparison for one- to eight-dimensional GPs after $10d$ evaluations, where $d$ is the dimension of the test function.}
\begin{tabular}{c@{\hspace{5pt}}l@{\hspace{5pt}}ccc}
\hline
Algorithm & Metric & \tt{best} & \tt{last} & \tt{two new} \\ 
\hline
EUBO & Perf. $\uparrow$ & $0.85 \pm 0.11$ & $0.83 \pm 0.12$ & $0.83 \pm 0.13$ \\
CrashPBO & Perf. $\uparrow$ & $\color{blue}{0.87 \pm 0.11}$ & $0.84 \pm 0.13$ & $0.85 \pm 0.11$ \\ 
\hline
EUBO & Crashes $\downarrow$ & $0.30 \pm 0.10$ & $0.34 \pm 0.14$ & $0.18 \pm 0.07$ \\ 
CrashPBO & Crashes $\downarrow$ & $\color{blue}{0.11 \pm 0.04}$ & $0.17 \pm 0.08$ & $0.12 \pm 0.05$ \\ 
\hline
\end{tabular}

\label{tab:results}
\end{table}

\section{Hardware Experiments} \label{sec:hexperiment}

We evaluate CrashPBO on three robotic platforms for two to four-dimensional parameter tuning tasks, demonstrating its ease of application across different experimental setups and tasks.
 For all experiments, we use the \emph{compare-to-best} mode and identical algorithm hyperparameters. Feedback is provided through a web-based interface, allowing DMs to express preferences, report crashes, and repeat uncertain trials. Each experiment highlights a different use case for CrashPBO.

\subsection{Quadcopter Backflips}
\label{sec:quad}

\textbf{Purpose:} Demonstrate CrashPBO to yield different outcomes based on user preferences and in a setting with costly resets (actual crashes).
 
\textbf{Setup:} We tune two timing parameters, $T_0$ and $T_1$, of a geometric backflip controller for a Crazyflie 2.1 quadcopter~\cite{antal2022nonlinear}. These parameters control the ascent and recovery durations. A Vicon motion capture system records the trajectory for feedback. Three DMs each conducted 15 trials. The DMs are lab members with subjective preferences. All DMs start from the same initial parameters that were selected through prior experiments. 

\begin{figure}
    \centering
    \includegraphics[width=1.0\linewidth]{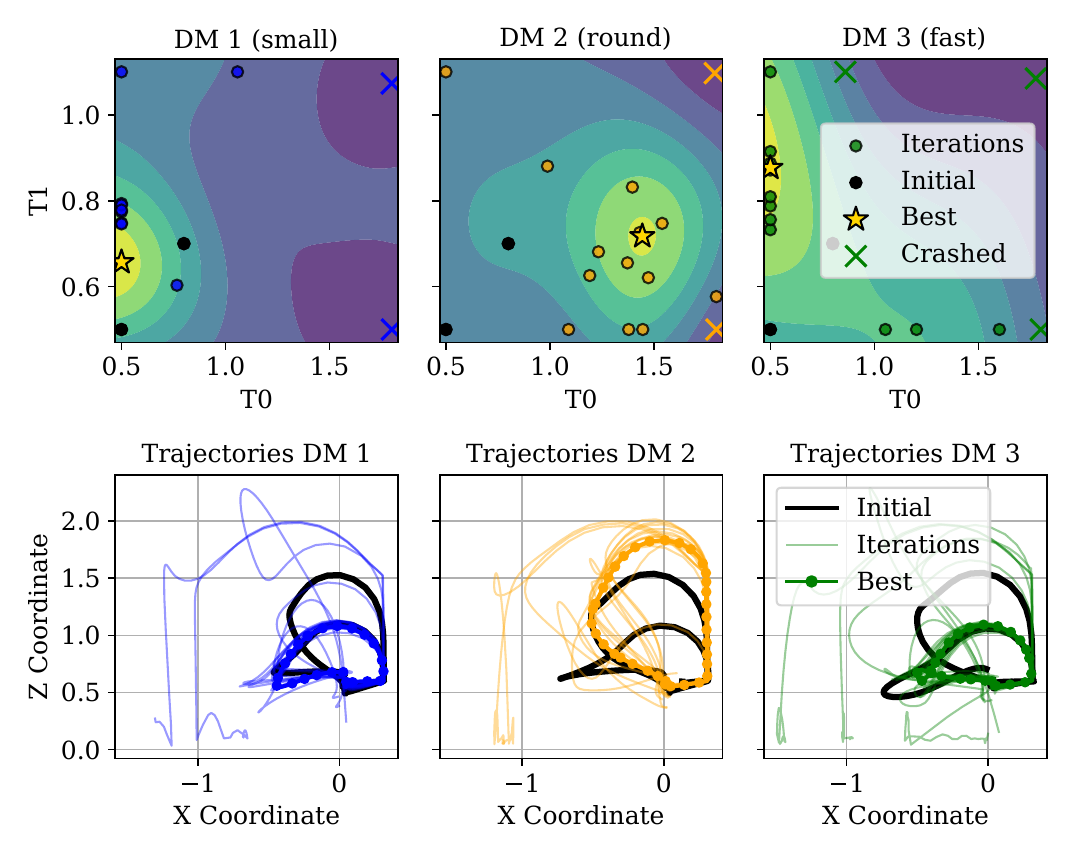}
    
    \caption{Experimental outcomes of three DMs tuning the backflip maneuver with subjective preferences. The top row shows the mean of the learned pairwise GP after the optimization, and the bottom row shows the trajectories of all evaluated flips. The black points indicate the shared initial parameters, which correspond to the black trajectories below. The final preferred parameters are marked with a star and appear as the bold trajectory. All other evaluated parameters are shown as points when the flip succeeded and as crosses when the experiment crashed.}
    \label{fig:backflip-results}
\end{figure}

\textbf{Results:} Figure~\ref{fig:backflip-results} shows the learned preference functions and resulting trajectories. Each DM converged to a distinct preferred flip without explicit objective definitions. DM~1 prefers a small backflip, DM~2 a round backflip, and DM~3 a fast backflip. Crashes occurred near parameter-space boundaries (two to three per DM) and required resets. The median duration is 75~s for crashed experiments and 25~s for non-crashed experiments. The mean durations are 77~s and 34~s, respectively. These numbers include long experiments for battery changes for the non-crashed experiments. Crash feedback prevented repeated sampling in unsafe regions, enabling efficient and personalized tuning.
\subsection{Furuta Pendulum}

\textbf{Purpose:} Demonstrate human-guided controller tuning and compare CrashPBO to baseline algorithms.

\textbf{Setup:} We use a Quanser Qube Servo 2 pendulum~\cite{furuta1992swing} to tune three parameters: two swing-up gains and the switching angle to an LQR controller. Two configurations are tested: one without and one with an added 5~g weight, starting from the same randomly sampled parameters, of which one is feasible. Each method (CrashPBO, EUBO, random search) is run for 20 iterations. The DM can repeat uncertain trials. One criterion for the DM is: Reduce the swing-up time, while the pendulum should not visit states with high $\theta$ values, as this leads to hitting against overswing protection, which causes wear and tear of the system. The DM can decide how conservative constraints are, and they do not need to be formulated directly. Code follows~\cite{bleher2022learning}. 

\begin{figure}
     \centering
     \begin{subfigure}[b]{0.12\textwidth}
         \centering
         \includegraphics[width=\textwidth]{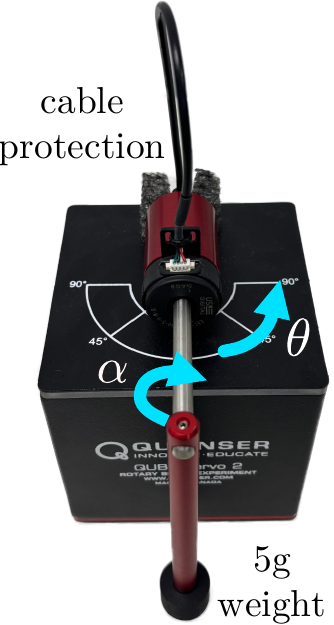}
         \caption{Pendulum}
         \label{fig:y equals x}
     \end{subfigure}
     \hfill
     \begin{subfigure}[b]{0.175\textwidth}
         \centering
         \includegraphics[width=\textwidth]{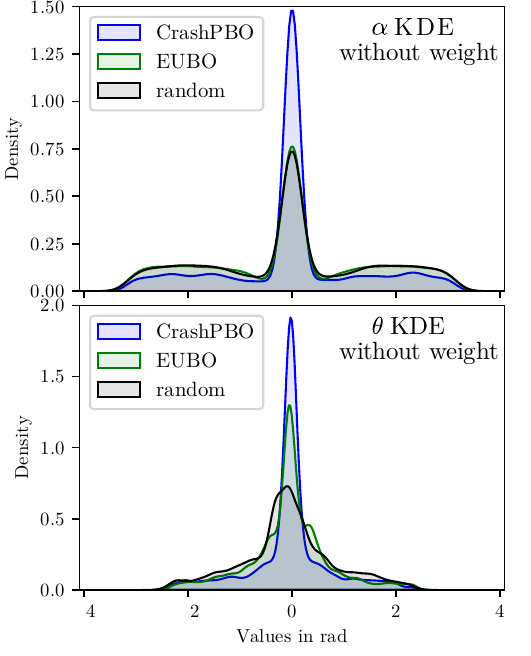}
         \caption{without weight}
         \label{fig:without weight}
     \end{subfigure}
     \hfill
     \begin{subfigure}[b]{0.175\textwidth}
         \centering
         \includegraphics[width=\textwidth]{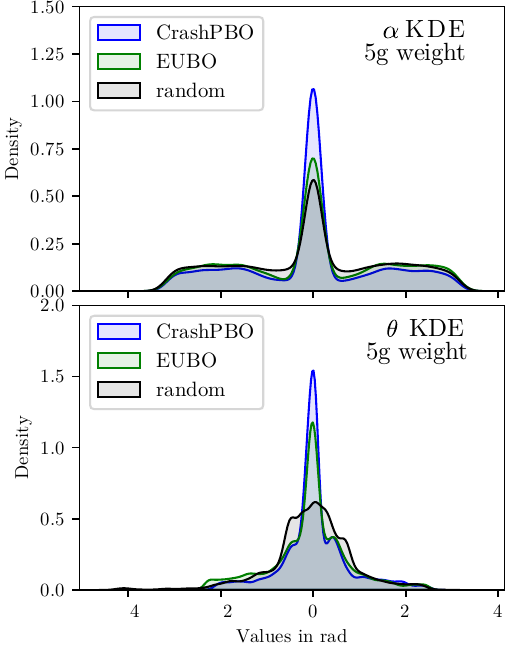}
         \caption{with added weight}
         \label{fig:with weight}
     \end{subfigure}
        
        \caption{(a) Furuta pendulum. (b)–(c) State distributions during learning with CrashPBO, EUBO, and random search, estimated using kernel density estimation (KDE). Successful swing-ups yield $\alpha,\theta \approx 0$, while large $\theta$ values indicate aggressive motion or cable protection impacts. CrashPBO’s lighter tails reflect more stable and less aggressive behavior. }
        \label{fig:quanserresults}
\end{figure}

\textbf{Results:} A feasible initial point was found after two random samples. One of the initial parameters resulted in a failed swing-up under CrashPBO but not under EUBO in the non-weight setting, highlighting outcome variability due to noise and differing initial states. Random search completed 20 iterations in about 5\,min, whereas human-in-the-loop methods required about 15\,min.
\begin{table}[ht]
\centering
\caption{Failed pendulum swing-ups}
\begin{tabular}{lrrr}
\hline
 & CrashPBO & EUBO & Random \\ 
\hline
no weights & 4/20 & 8/20 & 9/20 \\ 
weights & 6/20 & 8/20 & 12/20\\ 
\hline
\end{tabular}

\label{tab:resultspendulum}
\end{table}

The DM reported eight crashes in total (four failed swing-ups and four overly aggressive behaviors) in the setting without weights, and six in the setting with weights (all failed pendulum swing-ups). Table~\ref{tab:resultspendulum} summarizes unsuccessful swing-ups and shows that CrashPBO reduces these relative to the baselines. Figures~\ref{fig:without weight} and~\ref{fig:with weight} show the state distributions for $\alpha$ and $\theta$: CrashPBO yields sharper peaks at $\alpha = 0$ and lighter tails in $\theta$, indicating more successful swing-ups and better avoidance of large, wear-inducing angles, consistent with the optimization goal. The behavior changes noticeably when weight is added, reflecting the altered dynamics.

\subsection{Mini Wheelbot}

\textbf{Purpose:} CrashPBO to tune for subjective driving experience, while crashes act as strong rejections.

\textbf{Setup:} The Mini Wheelbot~\cite{hose2025mini} is a self-balancing unicycle robot, which receives manual steering commands in this experiment. We tune four parameters, \mbox{$\mathbf{x}=[x_{\mathrm{yaw}}, x_{\mathrm{drive}}, x_{\mathrm{friction}}, x_{\mathrm{slope}}]$}, affecting joystick scaling and low-level controller behavior. Two tasks are considered: ramp traversal and a figure-eight maneuver around cones. Each task consists of 20 experiments with up to five repetitions. The DM can report preferences or crashes and terminate runs freely. The initial parameters are chosen relatively conservatively for the joystick scaling and based on engineering knowledge from \cite{hose2024fine} for the low-level controller parameters.

\textbf{Results:} Table~\ref{tab:wheelbot_results} summarizes the outcomes. In the figure-eight task, four strong rejections occurred at domain edges due to unstable motion; in the ramp task, three crashes were reported for overly slow turning. Preferred parameters concentrate near mid-range input scalings, while controller parameters vary more widely, indicating lower sensitivity. Experiment durations range from 3~s to 286~s (mean 64~s), illustrating that CrashPBO supports non-episodic tuning.  

\begin{table}[]
    \centering
        \caption{Summary of Mini Wheelbot results, as in this setting it is hard to recall the best observed parameters, the best parameters can be repeated, so that not every experiment results in a decision.}
    \begin{tabular}{lcccc}
    \hline
    Experiment  & Decisions & Crashes & Final Parameters \\
    \hline
    Figure 8  & 15/20 & 4/20 & 0.79, 0.64, 0.0081, 0.82 \\
    Ramp  & 15/20 & 3/20 & 0.55, 0.55, 0.000, 2.61 \\
    \hline
    \end{tabular}

    \label{tab:wheelbot_results}
\end{table}
\subsection{Summary}

Our hardware results demonstrate that we can utilize PBO for user-friendly, personalized tuning in both subjective and objective tasks across episodic and non-episodic settings. The crash feedback mechanism effectively steers the optimization away from unwanted parameters, reducing the need for extensive tuning of parameter bounds in PBO.

\section{Conclusion}
In this letter, we proposed CrashPBO, a mechanism to provide crash feedback for PBO. In synthetic experiments, we demonstrated the potential to reduce crashes and increase data efficiency. In three robotic experiments, we showed how PBO can be effectively utilized for robotic parameter tuning, particularly in settings where a human can assess performance. We demonstrated CrashPBO for automatic controller tuning without the need to quantify an objective or constraint function on a Furuta pendulum and subjective preference tuning in a crash-prone setting when tuning quadcopter backflips. Further, we enabled preferential tuning with varying episode lengths on the Mini Wheelbot. 
CrashPBO, a hyperparameter-free algorithm, combines user preferences with crash feedback, circumventing tedious objective function tuning, making it a ready-to-use tool for robotic systems. While our DMs provided consistent feedback, future work should assess how feedback variability and quality affect CrashPBO’s robustness and include a ``cannot decide'' option for uncertain comparisons.

\section*{Acknowledgment}
The authors thank Alexander Gräfe, Paul Brunzema, Henrik Hose, Devdutt Subhasish, Sebastian Giedyk, and Miriam Kober for fruitful discussions and help with hardware setups.
 
\AtNextBibliography{\footnotesize}
\printbibliography

\end{document}